\documentclass[10pt,letterpaper]{article}
\usepackage[top=1in,left=1in,right=1in,bottom=1in]{geometry}

\usepackage{amsmath,amssymb}
\usepackage{changepage}
\usepackage[utf8x]{inputenc}
\usepackage{textcomp,marvosym}
\usepackage{cite}
\usepackage{nameref,hyperref}
\usepackage{microtype}
\DisableLigatures[f]{encoding = *, family = * }
\usepackage[table]{xcolor}
\usepackage{array}
\newcolumntype{+}{!{\vrule width 2pt}}
\newlength\savedwidth

\usepackage[aboveskip=1pt,labelfont=bf,labelsep=period,justification=raggedright,singlelinecheck=off]{caption}

\makeatletter
\renewcommand{\@biblabel}[1]{\quad#1.}
\makeatother

\usepackage{graphicx}
\usepackage{amsthm}
\usepackage[linesnumbered,ruled]{algorithm2e}

\DeclareMathOperator*{\argmax}{arg\,max}
\DeclareMathOperator*{\mean}{mean}

\newtheorem{theorem}{Theorem}
\title{Learning overcomplete, low coherence dictionaries with linear inference}
\author{Jesse A. Livezey\textsuperscript{1,2*},
	Alejandro F. Bujan\textsuperscript{2},
	Friedrich T. Sommer\textsuperscript{2}}
\begin{document}
\maketitle

\begin{flushleft}

\bigskip
\textbf{1} Lawrence Berkeley National Laboratory, Berkeley, California, USA
\\
\textbf{2} Redwood Center for Theoretical Neuroscience, University of California, Berkeley, California, USA
\\
\bigskip

* jlivezey@lbl.gov

\end{flushleft}
\begin{abstract}
Finding overcomplete latent representations of data has applications in data analysis, signal processing, machine learning, theoretical neuroscience and many other fields. In an overcomplete representation, the number of latent features exceeds the data dimensionality, which is useful when the data is undersampled by the measurements (compressed sensing, information bottlenecks in neural systems) or composed from multiple complete sets of linear features, each spanning the data space. Independent Components Analysis (ICA) is a linear technique for learning sparse latent representations, which typically has a lower computational cost than sparse coding, its nonlinear, recurrent counterpart. While well suited for finding complete representations, we show that overcompleteness poses a challenge to existing ICA algorithms. Specifically, the coherence control in existing ICA algorithms, necessary to prevent the formation of duplicate dictionary features, is ill-suited in the overcomplete case. We show that in this case several existing ICA algorithms have undesirable global minima that maximize coherence. Further, by comparing ICA algorithms on synthetic data and natural images to the computationally more expensive sparse coding solution, we show that the coherence control biases the exploration of the data manifold, sometimes yielding suboptimal solutions. We provide a theoretical explanation of these failures and, based on the theory, propose improved overcomplete ICA algorithms. All told, this study contributes new insights into and methods for coherence control for linear ICA, some of which are applicable to many other, potentially nonlinear, unsupervised learning methods.
\end{abstract}

\section{Introduction}\label{intro}

Mining the statistical structure of data is a central topic of machine learning and also is a principle for computational models in neuroscience. A prominent class of such algorithms is dictionary learning,  which reveal a set of structural primitives in the data, the dictionary, and a corresponding latent representation, often regularized by sparsity. Here we consider dictionary learning algorithms of the type first proposed under the name Independent Components Analysis (ICA)~\cite{Comon1994,Bell1997}, that are computationally light-weight because the learned mappings between data and latent representation are linear in both directions. In this work, we focus on overcomplete dictionary learning~\cite{olshausen1997,Hyvarinen2005,Le2011}, the case when the dimension of the latent representation exceeds the dimension of the data and therefore the linear filters (dictionary) generating the data cannot all be mutually orthogonal.

Studying methods for learning overcomplete dictionaries is motivated from many application. In data analysis, overcomplete dictionaries become essential if data are either undersampled~\cite{hillar2015}, or have a sparse structure with respect to a combination of orthobases~\cite{donoho2003}. In neuroscience, dictionary learning has not only been proposed for data analysis~\cite{delorme2007,agarwal2014,hirayama2015}, but also as a computational model for understanding the formation of sensory representations~\cite{Bell1997,Olshausen1996,Klein2003,Smith2006,Rehn2007,Zylberberg2011,Carlson2012}. In such computational models of sensory learning, overcompleteness is important. First, it has been estimated from anatomical data that in primary sensory areas the number of neurons by far exceeds the number of afferent inputs~\cite{barlow1981,spoendlin1989,curcio1990,leuba1994,northern2002,deweese2005}. Further, it has been shown that dictionary learning forms more diverse sets of features when overcomplete, which more closely matches the diversity of receptive fields found in sensory cortex~\cite{Rehn2007,Carlson2012,Olshausen2013}. 

ICA is a technique for learning the underlying non-Gaussian and independent sources, $S$, in a dataset, $X$. ICA is formulated as a noiseless linear generative model:
\begin{equation}\label{lin_gen}
X_i=\sum_{j=1}^L A_{ij}S_j,  
\end{equation}
where $A\in \mathbb{R}^{D\times L}$ is referred to as the \textit{mixing matrix} wherein $D$ is the dimensionality of the data, $X$, and $L$ is the dimensionality of the sources, $S$. The goal of ICA is to find the \textit{unmixing matrix} $W\in \mathbb{R}^{L\times D}$ such that the sources can be recovered, $S_i=\sum_jW_{ij}X_j$ with $W=A^{-1}$. In the complete case ($D=L$) the mixing matrix can be inverted. The unmixing matrix $W$ can then be obtained by minimizing the negative log-likelihood of the model:
\begin{equation}\label{ica}
-\log P(X;W)=\sum_{i=1}^{M} \sum_{j=1}^{L}
g(\sum_kW_{jk}X_k^{(i)}) - M\log(\det(W))
\end{equation}
where $g(\cdot)$ specifies the shape of the negative log-prior of the latent variables $S$ and is usually a smooth version of the $L_1$ norm such as the $\log(\cosh(\cdot))$, $X^{(i)}$ is the $i$th element of the dataset, $X$, which has $M$ samples, and where the bases are constrained to have unit-norm. The log-determinant comes from the multivariate change of variables in the likelihood from $X$ to $S$, 
\begin{equation}\label{change_of_variables}
    P(X)=P(S)|\det \frac{dS}{dX}|=P(W\cdot X)|\det W|.
\end{equation}
If the data has been whitened, the unconstrained optimization (Eq~\ref{ica}) can be replaced by a constrained optimization where the second term in the cost function is replaced with the constraint $WW^T=I$~\cite{Hyvarinen1997}. 

In complete ICA, the log-determinant (or the identity constraint) will prevent multiple elements of the dictionary, $W$, from learning the same feature. In overcomplete ICA, the linear generative model (Eq~\ref{lin_gen}) cannot be inverted, and therefore, overcomplete versions of Eqs~\ref{ica} and~\ref{change_of_variables} cannot be derived. One alternative to maximum likelihood learning is to create an objective function by adding a new cost, $C(W)$, to the sparsity prior~\cite{Hyvarinen2002,Le2011}. The new unconstrained objective function becomes
\begin{equation}\label{loss}
\text{Objective}(W) = \lambda \sum_{i=1}^{M} \sum_{j=1}^{L}
g(\sum_kW_{jk}X_k^{(i)}) +C(W).
\end{equation}
The cost, $C(W)$, should be chosen to exert coherence control on the dictionary, that is, to prevent the co-alignment of the bases. The coherence of a dictionary is defined as the maximum absolute value of the off-diagonal elements of the Gram matrix of a unit-normalized dictionary~\cite{Davenport2011}, $W$,
\begin{equation}\label{coherence}
\text{coherence}(W)\equiv\max_{i\ne j}|\sum_kW_{ik}W_{jk}|=\max_{i\ne j}|\cos\theta_{ij}|.
\end{equation}
A dictionary with high coherence (near 1) will have duplicated or nearly duplicated bases. Score matching~\cite{Hyvarinen2005} is another alternative to maximum likelihood learning which can be used for ICA models.

Sparse coding~\cite{Olshausen1996} is a dictionary learning method which requires an iterative, computationally complex \textit{maximum a posteriori} estimation or posterior estimation step. However, unlike ICA, sparse coding extends naturally to the overcomplete setting without modification. During inference, latent features in overcomplete sparse coding models~\cite{Lewicki1999} have an explaining-away effect on each other which encourages them to not learn coherent solutions. Sparse coding methods which add additional coherence costs have also been proposed~\cite{Ramirez2009,Sigg2012}.

A number of methods for coherence control in overcomplete ICA have been proposed including a quasi-orthogonality constraint~\cite{Hyvarinen1999}, a reconstruction cost (equivalent to the $L_2$ cost in Eq~\ref{eq:l2} below)~\cite{Le2011}, and a Random Prior cost~\cite{Hyvarinen2002} (see Section~\ref{methods} for details). However, a systematic analysis of the properties of proposed overcomplete ICA methods and a comparison with methods that extend more naturally to overcomplete representations, for example, sparse coding, is still missing in the literature.

Our first theoretical result is that although the global minima of the $L_2$ cost has zero coherence for a complete basis, in the overcomplete case, it has global minima with maximum coherence. We introduce an analytic framework for evaluating different coherence control costs, and propose several new costs, which fix deficiencies in previous methods. Our first novel approach is the $L_4$ cost on the difference between the identity matrix and the Gram matrix of the bases. The second method is a cost which we call the \textit{Coulomb} cost because it is derived from the potential energy of a collection of charged particles bound to the surface of an $n$-sphere. We also propose modifications to previously proposed methods of coherence control which allows them to learn less coherent dictionaries.

In addition to controlling coherence, we show empirically that these costs will influence the entire distribution of the learned bases in an overcomplete dictionary. We investigate the coherence control costs on model recovery on a dataset with known structure and finally, evaluate the diversity of bases learned on a dataset of natural image patches.

\section{Results}
In this section we first prove that the $L_2$ cost has global minima with coherence $=1$. We then propose new coherence control costs and evaluate them on a synthetic dataset and natural images.
\subsection{The $L_2$ cost has high coherence global minima}\label{sec:l2}
Dictionary or representation learning methods often augment their cost functions with additional terms aimed at learning less coherent features~\cite{Ramirez2009, Le2011, Sigg2012} or making learning through optimization more efficient~\cite{Howard2008}. The $L_2$ cost~\cite{Ramirez2009, Le2011, Sigg2012}, defined for a unmixing matrix, $W$, as:
\begin{equation}\label{eq:l2}
C_{L_2}(W) = \sum_{ij}(\delta_{ij}-\sum_kW_{ik}W_{jk})^2=\sum_{ij}(\delta_{ij}-\cos\theta_{ij})^2,
\end{equation}
 has been used to augment dictionary learning methods motivated by the desire to learn more incoherent dictionaries~\cite{Strohmer2003, Davenport2011}. However, we show that minimizing the $L_2$ cost is a necessary but not sufficient condition for finding \emph{equiangular tight frames} (see Section~\ref{l2_cost} for details and definitions), a certain class of minimum coherence solutions. Indeed, we prove that the $L_2$ cost has global minima with maximum coherence. This implies that the $L_2$ cost and its related costs are not providing coherence control in overcomplete dictionaries.

For the $L_2$ cost, it can be shown that for integer overcompleteness, there exists a set of global minima in which the angle between many pairs of bases is exactly zero and the coherence is 1, the maximum attainable value. We prove the following theorem:

\begin{theorem}\label{thm:globalminimum}
Let $W_0\in \mathbb{R}^{D \times L}$ be an overcomplete unmixing matrix with data dimension $D$ and latent dimension $L=n\times D$, with $n> 1,\ \in \mathbb{Z}$ and unit-norm rows. There exist dictionaries, $W_0$, that are global minima of the $L_2$ cost with coherence $=1$.
\end{theorem}

This shows that the $L_2$ cost has global minima that have the exact property it was proposed to prevent (high coherence). The proof of this theorem also shows that, in the complete case ($n=1$), an orthonormal basis is a global minimum of the $L_2$ cost. We also prove that there are operators which transform the pathological solution (coherence $=1$) into non-pathological solutions (coherence $< 1$) to which the $L_2$ cost is invariant:

\begin{theorem}\label{thm:rotations}
There exist non-trivial continuous transformations: $\Phi$, on $W_0$ to which the $L_2$ cost is invariant. These transformed dictionaries, $W_0\Phi$, have coherence $\le 1$ and are global minima of the $L_2$ cost.
\end{theorem}

These transformations will be constructed as rotations on $D$-dimensional subsets of the dictionary elements and rotate the subsets with respect to the remaining elements. Appendices~\ref{proof1} and~\ref{proof2} contain the proofs of these theorems.

\begin{figure}[!htbp]
  \centering
   \includegraphics{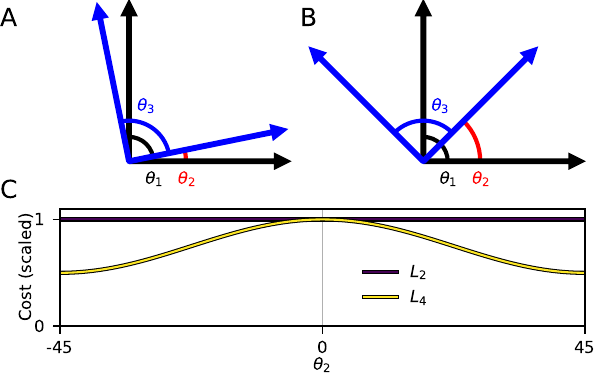}
  \caption{Structure of the pathological global minimum in the $L_2$ cost which the $L_4$ cost corrects. In \textbf{A} and \textbf{B}, each arrow represents a dictionary element in a 2-times overcomplete dictionary in a 2-dimensional space. \textbf{A} A dictionary with high coherence which has the same value of the cost as the dictionary in \textbf{B} for any $\theta_2$ including the pathological solution $\theta_2 \rightarrow 0$. \textbf{B} A dictionary with low coherence. \textbf{C} The $L_2$ and $L_4$ costs are plotted at $\theta_1 = \theta_3 = \pi/2$ as a function of $\theta_2$. The costs have been scaled so that their maximum value is 1.}
  \label{fig:pathological}
\end{figure}

These high coherence global minima are illustrated with a two dimensional, two times overcomplete example in Fig~\ref{fig:pathological}. It can be shown that there are pathological (high coherence) minima (Fig~\ref{fig:pathological}A) which can be continuously rotated into other low coherence minima (Fig~\ref{fig:pathological}B). These configurations are equivalent in terms of the value of the $L_2$ cost and lie on a connected global minimum. These families of configurations are minima if it can be shown that the gradient of the cost is zero, that is, they are critical points of the cost, and that the Hessian is positive definite in all directions but the one that rotates the configuration within the family of solutions. We will show these two things through an explicit derivation in the 2 dimensional case.

In order to understand these minima, we evaluate the $L_2$ cost in a two dimensional example analytically. The global rotational symmetry of the $L_2$ cost allows us to parameterize all solutions with respect to one fixed dictionary element: $( 1, 0 )$, without loss of generality. The four dictionary elements, shown in Fig~\ref{fig:pathological}, are:
\begin{equation}
(1,0),
(\cos\theta_1,\sin\theta_1),
(\cos\theta_2,\sin\theta_2),
(\cos\theta_2+\theta_3,\sin\theta_2+\theta_3).
\end{equation}
Setting $\theta_1$ and $\theta_3$ to $\pi/2$, that is, creating two sets of orthonormal bases, forms a ring of minima as $\theta_2$ is varied. This can be shown by computing the gradient and the eigenvalues of the Hessian of the cost at these points. The cost function, gradient, and Hessian are tabulated in Appendix~\ref{S1_Appendix} and the eigenvalues are plotted individually in Fig~\ref{fig:eig}.

The value of the $L_2$ cost is a constant as a function of $\theta_2$ (Fig~\ref{fig:pathological}C, purple line) even though the coherence is drastically changing as a function of $\theta_2$. These results show that the $L_2$ cost function does not provide coherence control. In fact, solutions that we would expect to be maxima are part of a set of global minima, indicating that there is a need for new forms of coherence control.

\subsection{Addressing high coherence solutions: $L_4$ and Coulomb costs}\label{sec:novel}

The rotational symmetry in the $L_2$ cost leads to its pathological (high coherence) global minima, and this insight motivates a simple modification which will not have high coherence minima. We propose a novel coherence control cost termed the $L_4$ cost, which transforms the pathological minima of the $L_2$ cost into saddle points. This cost function also acts on the gram matrix of $W$, but raises each off diagonal element to the fourth power which breaks the rotational symmetries which lead to the pathological minima
\begin{equation}
C_{L_4}(W) = \sum_{ij}(\delta_{ij}-\sum_kW_{ik}W_{jk})^4=\sum_{ij}(\delta_{ij}-\cos\theta_{ij})^4.
\end{equation}

Following the same analysis as in Section~\ref{sec:l2}, we show that the pathological solutions are either reduced to saddle points at $\theta_2=n\frac{\pi}{2}$ or local minima at $\theta_2=(2n+1)\frac{\pi}{4}$,  which correspond to incoherent solutions. The $L_4$ cost as a function of $\theta_2$ has a maximum at $\theta_2=0$ (coherent solutions) and minima at $\theta_2=\tfrac{\pi}{2}$ (Fig~\ref{fig:pathological}C). The $L_4$ cost function, gradient, and Hessian are tabulated in Appendix~\ref{S1_Appendix} for this 2 dimensional example.

We also propose a second alternative cost, where the repulsion from high coherence is \textit{Coulombic}: the Coulomb cost. Coherence control can then be related to the problem of characterizing the minimum potential energy states of $L$ charged particles on an $n$-sphere, an open problem in electrostatics~\cite{Smale1998}. The energy, $E^{\text{Coulomb}}$, of two charged point particles of the the same sign is proportional to the inverse of their distance, $\vec r_{ij}$:
\begin{equation}
E^{\textrm{Coulomb}}_{ij}\propto \frac{1}{|\vec r_{ij}|}.
\end{equation}
To map this problem onto ICA, the cost should be made symmetric around $\theta=\pi/2$ rather than $\theta=\pi$, which can be accomplished by replacing $\theta$ with $2\theta$, that is, $|r_{ij}|=\sqrt{1-\cos^2(\theta_{ij}/2)}\rightarrow\sqrt{1-\cos^2\theta_{ij}}$. Therefore, the Coulomb cost can be formulated as follows:
\begin{equation}
C_{\text{Coulomb}}(W)= \sum_{i\neq j}\frac{1}{\sqrt{1-\cos\theta_{ij}^2}}=\sum_{i\neq j}\frac{1}{\sqrt{1-\sum_kW_{ik}W_{jk}^2}}.
\end{equation}
In practice, we subtract the value of the cost for perpendicular bases, 1, for each pair $i \neq j$ to bring the cost into a better dynamic range. This cost diverges as coherence $\rightarrow 1$, which means it cannot have high coherence minima.

\subsection{Numerical investigations of coherence control}\label{analytic_numerical}

The above analysis provides evidence of a failure of the $L_2$ cost to provide coherence control. The alternative coherence cost function can prevent high coherence solutions, but all costs functions will act on the entire distribution of dictionary elements, not only the high coherence pairs. Deriving the distribution of pairwise angles in the minima of the cost functions is analytically difficult. However, understanding the influence of the coherence control cost function on the distribution of dictionary elements allows us to better understand their biases.

In order to understand the origin of the effects of the different coherence controls on the pairwise angle distributions, the coherence costs can be directly compared without the data dependent ICA sparsity prior. We use two different initializations of the bases and optimize the data-independent coherence costs. These initializations are: a noisy pathological initialization (as in Section~\ref{sec:l2}) and a random uniform initialization. We will numerically explore the minima of these cost function for a 2 times overcomplete dictionary in a 32 dimensional data space by minimizing the cost function with these two initializations.

\begin{figure}[!htbp]
  \centering
   \includegraphics{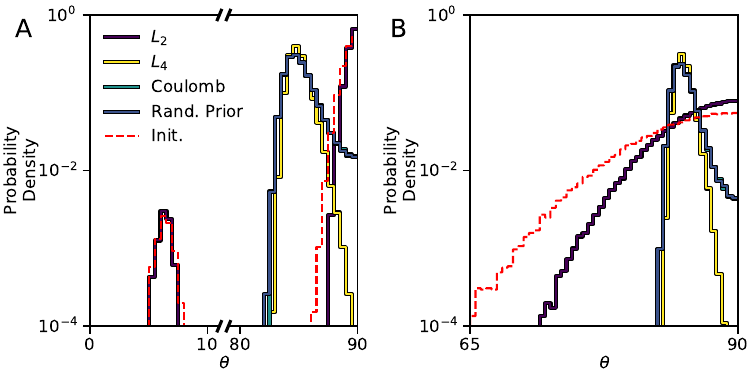}
  \caption{Coherence control costs have minima with varying coherence which can depend on initialization. Color legend is preserved across panels. For both panels a 2 times overcomplete dictionary with a data dimension of 32 was used and the distributions are averaged across 10 random initializations. \textbf{A} Distribution of pairwise angles (log scale) obtained by numerically minimizing a subset of the coherence cost functions for the pathological dictionary initialization. Red dotted line indicates the initial distribution of pairwise angles. Note that the horizontal axis is broken at 10 and 80 degrees. \textbf{B} Angle distributions obtained (as in \textbf{A}) from a uniform random dictionary initialization. Note that the horizontal axis only includes 65 to 90 degrees.}
  \label{fig:simulations}
\end{figure}

The noisy pathological initialization tiles an orthonormal, complete basis two times and adds a relatively small ($\sigma=.01$) amount of zero-mean Gaussian noise to every basis element to create $W$. As shown by the red-dashed histogram in Fig~\ref{fig:simulations}A, most pairwise angles start close to either 90 or 0 degrees as shown in the two peaks in the initial distribution. Minimizing the $L_2$ cost (purple line) from this initialization gives a final solutions with high coherence, similar to the initial distribution. The other costs push the pairs of bases with initially small pairwise angles apart. This shows numerically that the $L_2$ does not provide coherence control for overcomplete dictionaries unlike other proposed methods. Fig~\ref{fig:simulations_all} contains the same analysis for the full set of cost functions.

In the random uniform case, the elements of $W$ are drawn independently from a uniform distribution on the unit hyper-sphere. The final distribution of pairwise angles for the $L_2$ cost peaks at 90 degrees but also has a longer tail towards small pairwise angles. The other costs have shorter tails and have varying amounts of density near 90 degrees. Of all costs, the $L_4$ cost distributes the angles most evenly which is reflected by its distribution having the narrowest width and lowest coherence.

Together, these results show that the $L_2$ cost does not provide coherence control and is also sensitive to the initialization method. The proposed $L_4$ and Coulomb cost, as well as the previously proposed Random Prior (see Section~\ref{methods}), all provide coherence control. For these three costs, the distribution from which the dictionary was initialized does not have a large effect on the distributions at the numerical minima. These traits mean that they are better suited for providing coherence control in overcomplete dictionary learning methods.

\subsection{Flattened costs}

The previous analysis provides insight into why different cost function have different behavior for small angles (high coherence). However, the $L_4$, Coulomb, and Random Prior cost also show qualitatively different behavior in their distributions near 90 degrees. Both the Coulomb and Random Prior have density near 90 degrees for the distribution of pairwise angles, meaning that a fraction of the bases are nearly orthogonal. The $L_4$ has much lower density near 90 degrees, and a correspondingly lower coherence (smallest pairwise angle).

\begin{figure}[!htbp]
  \centering
   \includegraphics{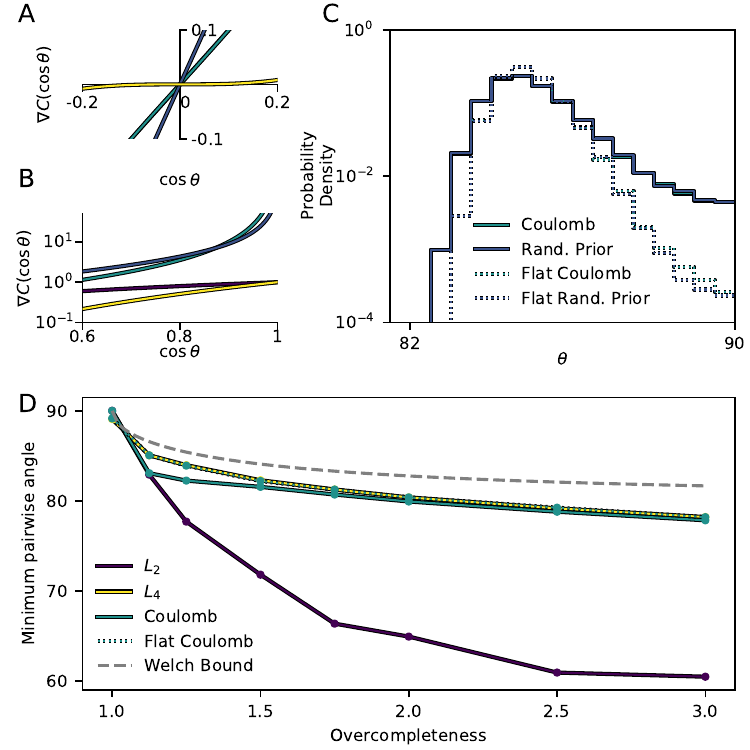}
  \caption{Quadratic terms dominate the minima of coherence control costs. \textbf{A} Gradient of the costs as a function of $\cos \theta$ near $\cos\theta = 0$. \textbf{B} Gradient of the costs as a function of $\cos\theta$ near $\cos\theta = 1$. \textbf{C} Distibution of pairwise angles for a 2 times overcomplete dictionary with a data dimension of 32 from 10 random uniform initializations. The Coulomb and Random Prior cost function distributions are shown (solid lines) along with their counterparts with quadratic terms removed (``flattened", dashed). \textbf{D} The median minimum pairwise angle (arccosine of coherence) across 10 initializations is plotted as a function of overcompleteness for a dictionary with a data dimension of 32. The largest possible value (Welch Bound) is also shown as a function of overcompleteness.}
  \label{fig:flat}
\end{figure}

In order to gain more insight into the causes of the qualitative differences in the distributions of angles, we analyze the behavior of the costs around $\theta=0$ and $\theta=90$ (Fig~\ref{fig:flat}A, B respectively). The gradient of the cost close to $|\cos\theta|=1$ is proportional to the force the angles feel to stay away from zero which will influence the high coherence tail of the distribution. Taylor expanding all the costs near $\cos\theta=0$ reveals that all cost functions have non-zero second order terms except for the $L_4$ cost which only has a fourth order term with linear and cubic terms in their gradients respectively as shown in Fig~\ref{fig:flat}A. Gradients which scale linearly will encourage pairs of basis vectors to be more orthogonal at the expense of skewing the angle distribution towards small values. This may lead to distributions of pairwise angles which are less uniform over all pairs of elements of the dictionary.

We hypothesize that the quadratic terms are creating higher coherence minima with more pairwise angles close to 90 degrees. This both motivates the $L_4$ cost and leads us to propose modified versions of the Coulomb and Random Prior costs where the quadratic terms have been removed. The Random Prior cost~\cite{Hyvarinen2002} is derived from the distribution of angles expected between pairs of angles randomly drawn on the surface of an $n$-sphere and is described in Section~\ref{methods}. This can be done by subtracting the quadratic term in the Taylor series from the original cost function,
\begin{equation}
C_\text{Flat}(\cos\theta_{ij}) = C(\cos\theta_{ij})-\left. \frac{\partial^2 C(\cos\theta_{ij})}{\partial \cos\theta_{ij}^2}\right|_0\cos^2\theta_{ij}.
\end{equation}

This hypothesis can be validated numerically. We compared the distribution of pairwise angles when the Coulomb nad Random Prior costs were minimized with their flattened counterparts. Both the Flattened Coulomb and Random Prior costs (Fig~\ref{fig:flat}C, dotted) show pairwise angle distributions which have lower coherence and fewer pairwise angles close to 90 degrees compared to the original costs (Fig~\ref{fig:flat}C, solid). This shows that across costs, the quadratic terms dominate the behavior of the pairwise angle distributions near 90 degrees and can have a small effect on the coherence on the distributions.

These coherence control methods will also have different behaviors as a function of overcompleteness. To understand their behavior, we measured the coherence of their minima as a function of overcompleteness. Fig~\ref{fig:flat}D shows the minimum pairwise angle ($\arccos$ of coherence, low coherence is high minimum pairwise angle) of these methods as a function of overcompleteness at fixed data dimensionality. The median over random initializations of the minimum pairwise angle between dictionary elements for numerically minimized coherence costs is shown. The cost functions evaluated here fall into three groups with quantitatively similar intra-group coherence as a function of overcompleteness. The $L_2$ cost has the highest coherence (smallest pairwise angle) for all overcompletenesses greater than 1. The $L_4$ cost and flattened versions of the Random Prior and Coulomb costs have the lowest coherence. The Random Prior and Coulomb costs behave similarly to the $L_2$ costs for low overcompleteness (less than 1.5) and then converge to be similar to the $L_4$ and flattened costs for high overcompletenesses (greater than 2). Fig~\ref{fig:3D_all} contains a detailed Coulomb and Random Prior comparison. The Welch Bound~\cite{welch1974} is the smallest possible coherence (largest minimum pairwise angle) achievable (Fig~\ref{fig:flat}D). The best coherence control cost functions approach, but do not saturate this bound. Note that constructing overcomplete dictionaries that saturate this bound for arbitrary overcompleteness is an open problem~\cite{Strohmer2003, fickus2015}. This shows that the quadratic terms in the cost function are dominating the coherence behavior of the cost functions and that removing the term as in the flattened costs or only including quartic terms as in the $L_4$ leads to lower coherence solutions.

These results show that proposed coherence control methods prevent high coherence to different degrees, and furthermore that the choice of coherence control, which is meant to affect the distribution of small pairwise angles, has an effect on the entire distribution of angles. Specifically, the $L_2$ cost does not provide coherence control and leads to solutions which are heavily biases by initialization unlike other proposed costs. These results also validate the relationship between second order terms in the cost function and the trade-off between coherence and orthogonality.

\subsection{Recovery of the mixing matrix with overcomplete ICA}\label{sec:recovery}

The previous analysis considered the data-independent coherence costs on their own. In ICA, the coherence costs will trade-off with the sparsity prior (Eq~\ref{loss}). Ideally, coherence costs would only prevent duplication of learned dictionary elements, but otherwise let the data shaping of the basis functions through the sparsity prior. In practice, we have shown that coherence control costs can have an effect on all dictionary elements, including those with large pairwise angles. It is not currently clear how these different costs will bias the learned dictionaries.

To investigate how the coherence control costs perform on data in overcomplete ICA, we compare different ICA cost functions and a sparse coding model on the task of recovering a known mixing matrix from $k$-sparse data with a Laplacian prior. We compare three classes of overcomplete dictionary recovery methods. The first is a sparse coding baseline~\cite{olshausen1997}, the second are maximum-likelihood inspired ICA models described in Section~\ref{intro} which combine the sparse prior from complete ICA and a coherence control cost, and the final is Score Matching~\cite{Hyvarinen2005}, which is a non-maximum-likelihood method that can be used in overcomplete ICA.

Overcomplete mixing matrices were generated from the Soft Coherence Cost (see Section~\ref{methods}) and used to generate a $k$-sparse dataset. The dictionary learning methods were then all trained on these datasets. Recovered unmixing matrices were compared to the ground-truth mixing matrix where the error for recovery is 0 for a perfect recovery ($W^T=A$) and 1 for a random recovery (see Section~\ref{metrics} for details). For a 32-dimensional data space, we vary the $k$-sparseness and overcompleteness of the data. For each of these datasets, where the number of dataset samples was 10-times the mixing matrix dimensionality, we fit all models to the data from 10 random initializations, for a range of sparsity weights: $\lambda$, if applicable, and then compare the recovery metric across models.

\begin{figure}[!htbp]
  \centering
   \includegraphics{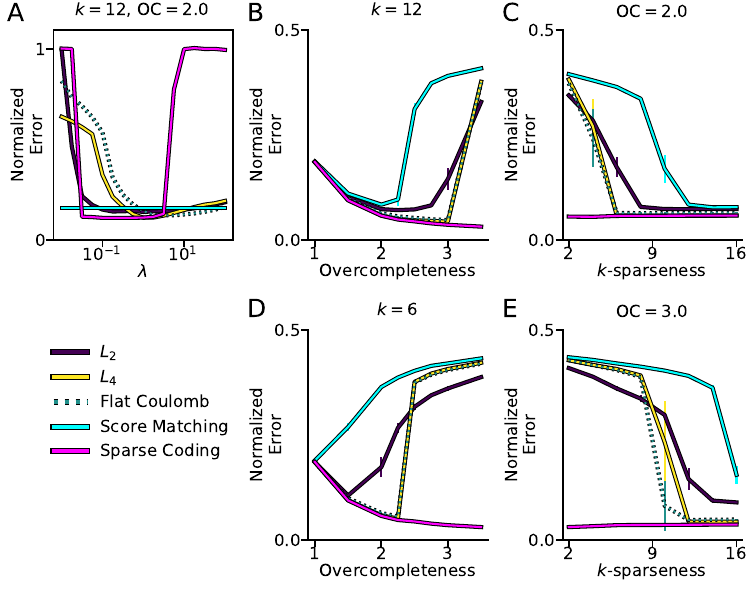}
  \caption{Coherence control costs do not all recover mixing matrices well. All ground truth mixing matrices were generated from the Soft Coherence cost and had a data dimension of 32. Color and line style legend are preserved across panels. \textbf{A} The normalized recovery error (see Section~\ref{methods} for details) for a 2-times overcomplete mixing matrix and $k=12$ as a function of the sparsity prior weight ($\lambda$).  Since score matching does not have a $\lambda$ parameter, it is plotted at a constant. \textbf{B} Recovery performance ($\pm$ s.e.m., $n=10$) at the best value of $\lambda$ as a function of overcompleteness at $k=12$. \textbf{C} Recovery performance ($\pm$ s.e.m., $n=10$) at the best value of $\lambda$ as a function of $k$-sparseness at 2-times overcompleteness. \textbf{D, E} Same plots as \textbf{B} and \textbf{C} at a point where methods do not perform as well: $k=6$ and 3-times overcomplete.}
  \label{fig:recovery}
\end{figure}

For a 12-sparse, 2-times overcomplete dataset, all methods can recover the mixing matrix well for some value of $\lambda$ (Fig~\ref{fig:recovery}A). The $L_2$ and Score Matching costs perform slightly worse than the maximum-likelihood inspired ICA methods and sparse coding. All methods have a certain range of $\lambda$ over which they recover the mixing matrix well and have differences in how they fail, for instance sparse coding has a very quick transition to poor recovery compared to ICA methods whose performance tends to decrease more slowly as $\lambda$ moves outside of the optimal range.

At fixed $k$-sparsity ($k=12$), we vary the overcompleteness and compare recovery costs (Fig~\ref{fig:recovery}B). As a function of overcompleteness, Score Matching recovers well in a smaller range of overcompleteness as compared to other ICA methods. Besides the $L_2$ cost, all other ICA methods have nearly identical recovery. The $L_2$ cost's performance breaks down at lower overcompleteness. All ICA methods fail to recover the mixing matrix once the overcompleteness becomes too large, while sparse coding continues to succeed in recovering the mixing matrix. Since the number of bases being recovered changes as the overcompleteness changes, it is not meaningful to compare the recovery metric between overcompletenesses, but it meaningful to compare different models at fixed overcompleteness.

At fixed overcompletenesss (OC=2), we vary the $k$-sparsity and compare recovery costs Fig (\ref{fig:recovery}C). Sparse coding performs well at all $k$-sparsenesses, but the ICA methods perform better with larger $k$-sparseness. The $L_2$ cost and Score Matching fails to recover well at a lower $k$-sparseness than other ICA methods. Since the number of bases being recovered is fixed as a function of the $k$-sparseness, the recovery metric can be compared across $k$-sparseness and models.

Fig~\ref{fig:recovery}D and E show the methods in a regime ($k=6$ and 3-times overcomplete, respectively) where ICA methods do not recover the mixing matrix as well as sparse coding. Fig~\ref{fig:recovery_all} contains the same analysis for the full set of cost functions.

In summary, we find different ICA methods have different regimes of performance with Score Matching and the $L_2$ cost having the smallest ranges of applicability. Other ICA methods generally have similar performance. Score Matching did not always perform as well as other ICA methods as a function of overcompleteness or $k$-sparseness, although it is a hyperparameter-free cost (no $\lambda$ hyperparameter). In all cases, the more computationally costly sparse coding was able to recover the mixing matrix more consistently than ICA models. This suggests that the linear inference in ICA models can only recover dictionaries for moderately overcomplete representations.

\subsection{Experiments on natural images}\label{nat_images}

When ICA is applied to real data, one typically does not know the exact generative distribution of the data. For instance, for a natural images dataset, we no longer have a ground truth mixing matrix or known prior, and furthermore, it is not likely that natural image patches come from a simple ICA-like generative model~\cite{hyvarinen2007, lucke2009}. However, the effects of coherence control on the distribution of dictionary elements learned can be evaluated. Specifically, we can look at the coherence of learned dictionaries and whether different methods prevent duplicate features from being learned.

We train 2-times overcomplete ICA models on 8-by-8 whitened image patches from the Van Hateren database~\cite{vanhateren1998} at a fixed value of sparsity across costs found by binary search on $\lambda$. The score matching cost has no $\lambda$ parameter to trade off sparsity versus coherence although it finds solutions of similar sparsity to the value chosen for the other costs. It is known that for natural images data sets, bases learned with ICA can be well-fit by Gabor filters~\cite{Bell1997}. Hence, we evaluate the distribution of the learned basis by inspecting the parameters obtained from fitting the bases to Gabor filters (see Section~\ref{gaborfit} for details).

\begin{figure}[!htbp]
  \centering
   \includegraphics{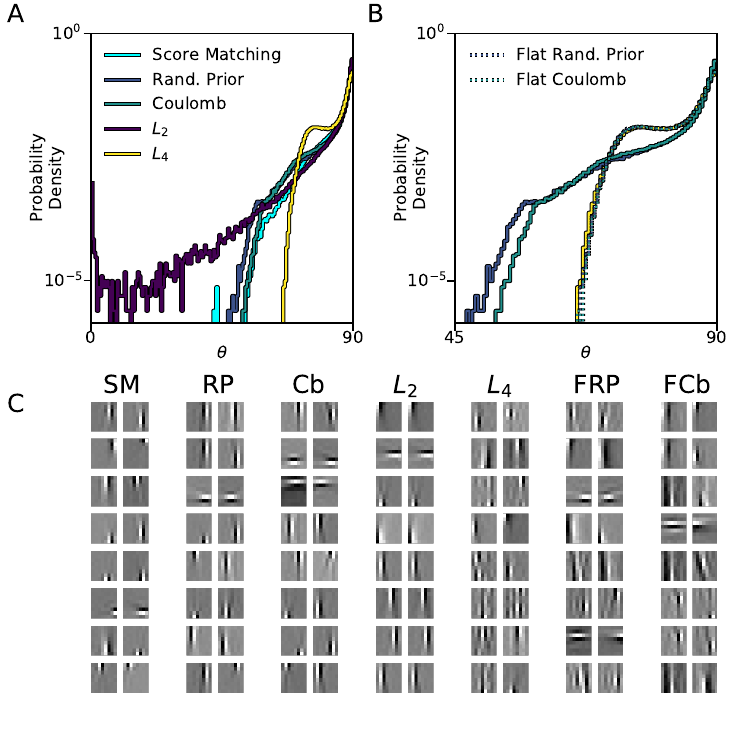}
  \caption{The coherence of an overcomplete dictionary learned from natural images depends on the coherence control cost. Results from fitting a 2-times overcomplete model on 8-by-8 natural image patches. \textbf{A, B} Pairwise angle distributions (log scale) across costs for the learned dictionaries for a fixed value of sparsity across costs. \textbf{B} Comparison between the Random Prior and Coulomb costs and their flattened versions. The $L_4$ distribution is also shown for comparison. Note that the horizontal axis covers 45 to 90 degrees. \textbf{C} For each cost from \textbf{A} and \textbf{B}, the 8 pairs of bases with smallest pairwise angle are shown. Since the overall sign of a basis element is arbitrary, the bases have been inverted to have positive inner product, if needed, for visualization.}
  \label{fig:naturalscenes1}
\end{figure}

The distributions of angles from the trained ICA models are in line with the theoretical results from Section~\ref{analytic_numerical}. The $L_2$ cost has more pairwise angles close to zero compared to the other costs with the $L_4$ having the smallest coherence. Similarly, as shown in Fig~\ref{fig:naturalscenes1}B, the Random Prior and Coulomb costs have lower coherence when the second order terms are removed and behave more similarly to the $L_4$ cost. These distributions also show that ICA models with the $L_2$ cost tend to learn duplicate bases from natural images.

\begin{figure}[!htbp]
  \centering
   \includegraphics{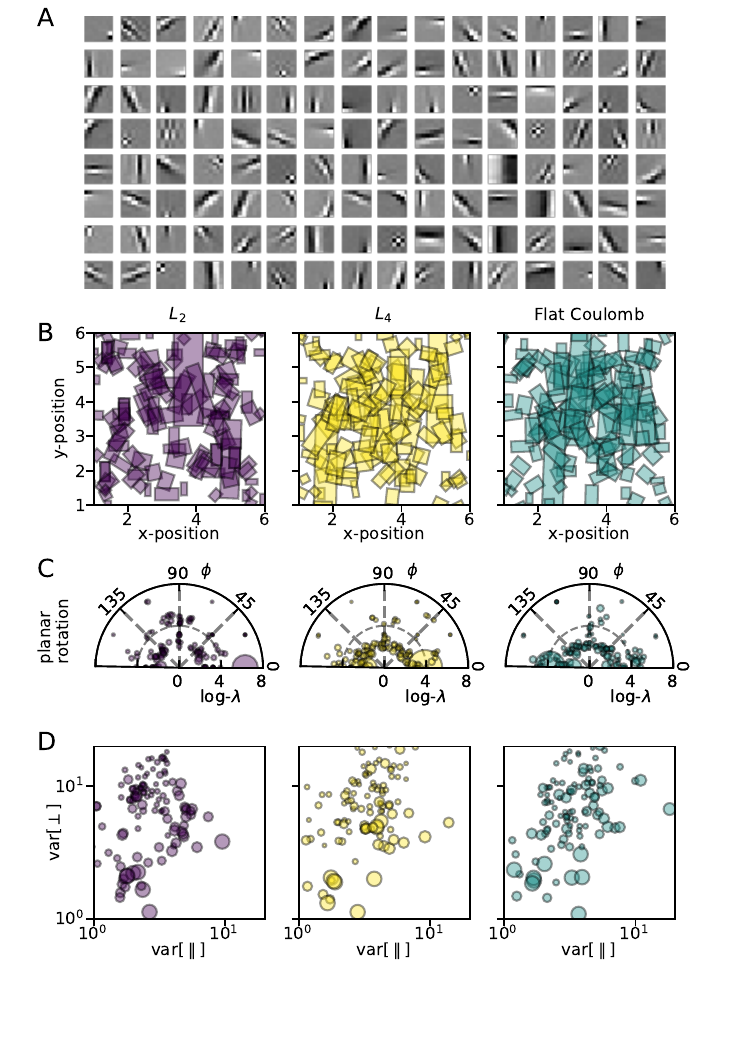}
  \caption{All coherence costs learn a dictionary that approximately tiles the space of Gabor Filters. \textbf{A} Dictionary learned using the $L_4$ cost on 8-by-8 natural image patches. \textbf{B} Distributions of locations, envelope scales, and rotations. Rectangle position: center of Gabor fit in pixel coordinates, rectangle rotation: planar-rotation of the Gabors, rectangle shape: envelope width parallel and perpendicular to the oscillation axis. \textbf{C} Distributions of rotations, log-wavelengths ($\lambda$), and envelope widths. Polar plots of planar-rotation angle and log-spatial wavelength of the Gabors. Marker size scales with geometric mean of envelope widths. \textbf{D} Distributions of envelope scales and log-wavelengths. Log-scale plot of envelope widths-squared parallel and perpendicular to the oscillation axis of the Gabors. Circle size scales with log-wavelength.}
  \label{fig:naturalscenes2}
\end{figure}

For the range of sparsities which were considered, the visual appearance of the individual bases is similar to results from previous ICA work and similar across costs ($L_4$ bases are shown in Fig~\ref{fig:naturalscenes2}A). However, none of the ICA methods considered here approximate the distribution of simple cell receptive fields reported in the early visual system~\cite{Ringach2002, Rehn2007, Zylberberg2011} (see Fig~\ref{fig:v1_comparison} for comparison). The tiling properties of the learned dictionaries can also be visualized directly. The coordinates of the center of the fit Gabor filter, rotations, and scales tile the space for the $L_2$, $L_4$, and Flattened Coulomb costs (Fig~\ref{fig:naturalscenes2}B). The dimensions and rotation of the rectangle represent the envelope widths and planar rotation angle respectively. This is similarly true for the planar rotation angle against the oscillation wavelength of the Gabor (Fig~\ref{fig:naturalscenes2}C) and the envelope widths and wavelengths (Fig~\ref{fig:naturalscenes2}D). Although these distributions look qualitatively similar, the underlying dictionaries can have very different coherence.

These results demonstrate that the $L_2$ cost learns undesirable, high-coherence overcomplete dictionaries on real data. Visually inspecting the bases or even their tiling properties may not reveal the redundant set of basis functions. To reveal this type of redundancy one has to measure the coherence or the distribution of pairwise angles of a dictionary directly.

\section{Methods}\label{methods}

In this section we summarize previously proposed coherence control methods, our model implementations, and datasets used.

\subsection{Previously proposed coherence control methods}
\paragraph*{Reconstruction cost and the $L_2$ cost}\label{l2_cost}
Le et al.~\cite{Le2011} propose adding a reconstruction cost to the ICA prior (RICA) as a form of coherence control, which they show is equivalent to a cost on the $L_2$ norm of the difference between the Gram matrix of the filters and an identity matrix for whitened data
\begin{equation}
\begin{split}
C_{\text{RICA}} &= \frac{1}{N}\sum_{ij}(X^{(i)}_j-\sum_{kl}W_{kj}W_{kl}X^{(i)}_l)^2\\
\propto C_{L_2} &= \sum_{ij}(\delta_{ij}-\sum_kW_{ik}W_{jk})^2=\sum_{ij}(\delta_{ij}-\cos\theta_{ij})^2,
\end{split}
\end{equation}
where $W_{ij}$ is the component of the $i$th source for the $j$th mixture , $X^{(i)}_j$ is the $j$th element of the $i$th sample, $\theta_{ij}$ is the angle between pairs of basis, and $\delta_{ij}$ is the Kronecker delta.

The $L_2$ cost has also been proposed as a form of coherence control~\cite{Ramirez2009,Sigg2012}. Equiangular tight-frames (ETFs) are a set of frames (overcomplete dictionaries) which have minimum coherence. The fact that an ETF has minimum coherence is used to motivate the $L_2$ cost as a form of coherence control.
A matrix $W \in \mathbb{R}^{L\times D}$ is an ETF if
\begin{equation}\label{eq:equiangular}
\sum_k W_{ik}\cdot W_{jk}=\cos \alpha,\ \forall i \neq j
\end{equation}
for some angle, $\alpha$, and
\begin{equation}\label{eq:outer}
\sum_k W_{ki}W_{kj} = \frac{L}{D}\delta_{ij}.
\end{equation}
The $L_2$ cost will encourage Eq~\ref{eq:outer} to be satisfied, but does not encourage Eq~\ref{eq:equiangular} to be satisfied as we show in Theorem~\ref{thm:globalminimum}.

\paragraph*{Quasi-orthogonality constraint}
Hyv{\"a}rinen et al.~\cite{Hyvarinen1999} suggest a quasi-orthogonality update which approximates a symmetric Gram-Schmidt orthogonalization scheme for an overcomplete basis, $W$, which is formulated as:
\begin{equation}
W \leftarrow\frac{3}{2}W-\frac{1}{2}WW^TW.
\end{equation}

\paragraph*{Random prior cost}
A prior on the distribution of pairwise angles was proposed to encourage low coherence~\cite{Hyvarinen2002}. The prior is the distribution of pairwise angles for two vectors drawn from a uniform distribution on the $n$-sphere\footnote{For both the Random Prior and the Coulomb cost, we regularize the costs and their derivatives near $|\cos\theta|=1$ by adding a small positive constant in the objective: $1-\cos\theta_{ij}^2\rightarrow 1+|\epsilon|-\cos\theta_{ij}^2$.}.
\begin{equation}
C_\text{Random prior}=-\sum_{i\ne j}\log P(\cos\theta_{ij}) \propto -\sum_{i\ne j}\log(1-\cos^2\theta_{ij})
\end{equation}

\paragraph*{Score Matching}
Score matching is a training objective function for non-normalized statistical models of continuous variables\cite{Hyvarinen2005}. It has been used to learn overcomplete ICA models. The score function is derivative of the log-likelihood of the model or data distribution with respect to the data
\begin{equation}
\psi(X;\Theta) = \nabla_X \log p(X;\Theta)
\end{equation}
The score matching objective is the mean-squared error between the model score, $\psi(X;\Theta)$, and data score, $\psi_\mathcal{D}(X;\Theta)$ averaged over the data, $\mathcal{D}$:
\begin{equation}
J(\Theta) = \frac{1}{2}\int_X p_\mathcal{D}(X)||\psi(X;\Theta)-\psi_\mathcal{D}(X;\Theta)||^2.
\end{equation}

\subsection{Coherence-based costs}
The coherence of a dictionary is defined as the maximum absolute value of the off-diagonal elements of the Gram matrix~\cite{Davenport2011} as in Eq~\ref{coherence}, which can be used as a cost function during optimization. This cost is difficult to numerically optimize since the derivative through the $\max$ operation will only act on one pair of bases at each optimization step, although it should find solution with local minima of coherence. An easier to optimize, but heuristic, version of this cost is the sum over all off-diagonal elements whose squares are larger than the mean squared value
\begin{equation}
C_\text{Soft Coherence}=\sum_{i\neq j\ \text{s.t.}\ \cos\theta_{ij}^2 > \cos\hat\theta^2}|\cos\theta_{ij}|,\textrm{ with }\cos\hat\theta^2 = \mean_{i\neq j}(\cos\theta_{ij}^2).
\end{equation}
We find that this cost does not work well for coherence control in ICA when fit with data, but it can be used to create low-coherence mixing matrices for generating data with known structure in Section~\ref{sec:recovery}.

\subsection{Model implementation}
All models were implemented in Theano~\cite{Theano2016}. ICA models, with the exception of the Coherence cost, were trained using the L-BFGS-B~\cite{Byrd1995} implementation in SciPy~\cite{Jones2001}. FISTA~\cite{beck2009} was used for MAP inference in the sparse coding model and the weights were learned using L-BFGS-B. All weights were training with the norm-ball projection~\cite{Le2011} to keep the bases normalized. A repository with code to reproduce the results will be posted online. For ICA models with coherence costs, the coherence control cost with no sparsity penalty ($\lambda=0$) was used as the objective for Figs~\ref{fig:simulations} and~\ref{fig:flat}.

\subsection{Datasets}
For all datasets and models, the number of samples in a dataset was equal to 10 times the number of model parameters, that is, $10\times n_\text{sources}\times n_\text{mixtures}$.

\subsubsection{$k$-sparse datasets}
Mixing matrices were generated by minimizing the Soft Coherence cost. Data was generated by keeping $k$ elements from a diagonal multivariate Laplacian distribution, zeroing out the rest, and combining them with the mixing matrix. 

\subsubsection{Natural images dataset}
Images were taken from the Van Hateren database~\cite{vanhateren1998}. We selected images where there was no evident motion blur and minimal saturated pixels. 8-by-8 patches were taken from these images and whitened using PCA.

\subsection{Dictionary recovery error}\label{metrics}
If the mixing matrix $A$ is recovered perfectly, $W^T$ will be a permutation of $A$. To estimate the closeness to a permutation matrix, the matrix $P_{ij}=|A_i\cdot W^T_j|$ is created. The largest element of the matrix is found which correpsonds to some $i, j$. The $\arccos$ of this element (angle between $A_i$ and $W^T_j$) is taken and added to a list and then the dictionary elements $A_i, W^T_j$ are removed. This processes is repeated until there are no more dictionary elements and then the median of the angles is returned as the error. The pseudocode for this algorithm is shown in Algorithm~\ref{alg:error}.
\begin{algorithm}
    \SetKwInOut{Input}{Input}
    \SetKwInOut{Output}{Output}

    \underline{function ERROR} $(A,W)$\;
    \Input{A ground truth mixing matrix, $A$, and recovered unmixing matrix, $W$}
    \Output{Median recovery angle error}
    $arr = \text{list}()$\;
    \For{$n=1$ \KwTo $n_\text{sources}$}{
        $i,j=\argmax_{n,m}{|A_n\cdot W^T_m|}$\;
        $arr.\text{append}(\arccos(|A_i\cdot W^T_j|))$\;
        $\text{del}\ A_i,\ \text{del}\ W^T_j$\;
    }
    \Return median($arr$)
    \caption{Algorithm for computing dictionary recovery error}
    \label{alg:error}
\end{algorithm}

This error is normalized by calculating the same quantity for matrices, $W^8$, which were recovered from mixing matrices $A^*$, which were from the same distribution as $A$ but with different random initializations. After this normalization, perfect recovery gives a normalized error of 0 and a random recovery gives a normalized error of 1.

\subsection{Fitting Gabor parameters}\label{gaborfit}
We fit the Gabor parameters~\cite{Ringach2002} to the learned bases using an iterative grid-search and optimization scheme which gave the best results on generated filters. The learned parameters were the center vector: $\{\mu_x, \mu_y\}$, planar-rotation angle: $\theta$, phase: $\phi$, oscillation wave-vector $k$, and envelope variances parallel and perpendicular to the oscillations: $\sigma_{\hat x}^2$ and $\sigma_{\hat y}^2$ respectively. Because they are constrained to be positive, the log of the parameters: $\sigma_{\hat x}^2$ and $\sigma_{\hat y}^2$ are optimized. To keep the wavelength of the Gabor larger than $2\sqrt{2}$ pixels, instead of optimizing $k$ directly we optimize $\rho$ with $k=\frac{2\pi}{2\sqrt{2}+\exp(\rho)}$. Shorter wavelengths are aliased by the pixel sampling.

\begin{equation}
\begin{split}
\hat x & = \cos(\theta)x + \sin(\theta)y\\
\hat y & = -\sin(\theta)x + \cos(\theta)y\\
\hat \mu_x & = \cos(\theta)\mu_x + \sin(\theta)\mu_y\\
\hat \mu_y & = -\sin(\theta)\mu_x + \cos(\theta)\mu_y\\
\text{Gabor}(x, y;\mu_x, \mu_y, \theta, \sigma_{\hat x}, k, \sigma_{\hat y}, \phi) &= \exp\left(-\frac{(\hat x - \hat \mu_x)^2}{2\sigma_{\hat x}^2}-\frac{(\hat y - \hat \mu_y)^2}{2\sigma_{\hat y}^2}\right)\sin(k \hat x+\phi)
\end{split}
\end{equation}

The procedure for finding the best Gabor kernel parameters was to save the parameter set with best mean-squared error after the following iterations:
\begin{enumerate}
	\item for different initial envelope widths, fit the center location for the envelope to the blurred absolute value of the basis,
	\item for different initial planar rotations and frequencies, numerically optimize the rotation, phase, and frequency of the Gabor
	\item for the best fit from above, re-optimize the centers, widths, and phases,
    \item re-optimize all parameters from best previous fit.
\end{enumerate}

A repository with code to fit the Gabor kernels is posted online \footnote{\url{https://github.com/JesseLivezey/gabor_fit}}.

\section{Discussion}

Learning overcomplete sparse representations of data is often an extremely informative first stage in analyzing multivariate data, such as sensor and measurement data. In the field of neuroscience, sparse coding serves not only a method for analyzing experimental data~\cite{agarwal2014}, but also as a computational model of how the brain analyzes sensory inputs~\cite{Olshausen1996, Rehn2007, Smith2006}. For all these purposes, the heavy computational cost of the nonlinear inference step involved in common sparse coding approaches is a major obstacle. It slows the analysis of large data sets, and also poses questions whether computational models for sensory systems with such high computational demands are compatible with the speed and ease of perception behaviors. For learning complete sparse representations, ICA with just a linear inference mechanism is a viable alternative with drastically reduced computational demand. Here, we investigated potential and limitations of linear inference methods in overcomplete dictionary learning. 

Any multidimensional method for extracting signal components needs a form of coherence control to prevent components from becoming co-aligned and therefore redundant. We first compared different coherence costs' ability to prevent the learning of coherent dictionary elements in the overcomplete case. We show theoretically and by simulation, that the $L_2$ cost, which successfully achieves orthogonality in the complete case, exhibits pathological global minima with maximum coherence in the overcomplete case.

We then suggest novel cost functions which do not suffer from pathological minima in the overcomplete case. Specifically, we propose the $L_4$ cost and the flattened versions of the Coulomb and Random Prior costs, and show that they yield dictionaries with lower coherence than the cost functions that have been proposed earlier. At the same time, these new cost functions have smaller effects on incoherent basis pairs, thus leading to dictionaries that reflect the structure of the data rather than effects from the coherence term.

Further, we show that the methods of coherence control proposed here can successfully learn representations with linear inference in certain regimes of overcompleteness and sparseness, in which standard ICA methods fail. However, this expansion of the regime of applicability is still limited. Even the improved methods begin to fail when overcompleteness grows beyond two-fold (for 32 dimensional data) or if the data is $k$-sparse with small $k$. The problem to deal with extremely $k$-sparse data is counterintuitive at first, because nonlinear inference methods usually do better as $k$ is decreased because the combinatorial search for the best sparse support in the inference becomes easier~\cite{Davenport2011}. However, linear inference in ICA models cannot recover extremely sparse sources unlike sparse coding models, which do not fail in the small-$k$ limit.

All told, our study explores the power and limitations of linear inference for overcomplete dictionary learning. We note that variations of the ICA sparsity prior and mismatch with data sparsity structure have not been systematically explored here and are another potential topic of further investigation. The limitations of linear methods to yield highly sparse, highly overcomplete representations might suggest a reason why cortex provides dense local recurrent networks in early sensory areas. The circuitry could provide the substrate for nonlinear inference of sparse sensory representations that possess an overcompleteness which has been estimated to be, depending on species, between ten- and many hundred-fold~\cite{spoendlin1989,curcio1990,leuba1994,northern2002,deweese2005}.

\section*{Acknowledgments}
We thank Yubei Chen, Alexander Anderson, and Kristofer Bouchard for helpful discussions. JAL was supported by the Laboratory Directed Research and Development Program of Lawrence Berkeley National Laboratory under U.S. Department of Energy Contract No. DE-AC02-05CH11231. JAL and AFB were supported by the Applied Mathematics Program within the Office of Science Advanced Scientific Computing Research of the U.S. Department of Energy under contract No. DE-AC02-05CH11231. FTS was supported by the National Science Foundation grants IIS1718991, IIS1516527, INTEL, and the Kavli Foundation. We acknowledge the support of NVIDIA Corporation with the donation of the Tesla K40 GPU used for this research. 

\bibliographystyle{unsrt}
\bibliography{references}

\newpage
\appendix

\section{Minima analysis for the $L_2$ and $L_4$ costs for a 2-dimensional space.}
\label{S1_Appendix}
\renewcommand{\thefigure}{A\arabic{figure}}

\setcounter{figure}{0}
Here we tabulate the full Hessian matrices, eigenvalues, and eigenvectors for the analysis in Sections~\ref{sec:l2} and~\ref{sec:novel}.

\subsection{$L_2$ cost}
\begin{equation}
\begin{split}
C_{L_2}(\theta_1, \theta_2, \theta_3)|_{\theta_1, \theta_3=\tfrac{\pi}{2}} &= 4 \\
\frac{\partial C_{L_2}(\theta_1, \theta_2, \theta_3)}{\partial \vec \theta}|_{\theta_1, \theta_3=\tfrac{\pi}{2}} &= \begin{pmatrix}
0 & 0 & 0
  \end{pmatrix}\\
H(C_{L_2})|_{\theta_1, \theta_3=\tfrac{\pi}{2}} & =\begin{pmatrix}
4 & 0 & 4 \cos 2 \theta_2 \\
0 & 0 & 0 \\
4 \cos 2 \theta_2 & 0 & 4
  \end{pmatrix}\\
\text{EVal.}(H_{L_2})|_{\theta_1, \theta_3=\tfrac{\pi}{2}} & =\begin{pmatrix}
0 \\ 8\sin^2\theta_2 \\ 8\cos^2\theta_2
  \end{pmatrix}\\
\text{EVec.}(H_{L_2})|_{\theta_1, \theta_3=\tfrac{\pi}{2}} & =
\begin{pmatrix}
0 \\
1 \\
0 
 \end{pmatrix},
\begin{pmatrix}
-1 \\
0 \\
1 
 \end{pmatrix},
\begin{pmatrix}
1 \\
0 \\
1 
 \end{pmatrix}
\end{split}
\end{equation}

\subsection{$L_4$ cost}
\begin{equation}
\begin{split}
C_{L_4}(\theta_1, \theta_2, \theta_3)|_{\theta_1, \theta_3=\tfrac{\pi}{2}} &= 3+\cos 4 \theta_2 \\
\frac{\partial C_{L_4}(\theta_1, \theta_2, \theta_3)}{\partial \vec \theta}|_{\theta_1, \theta_3=\tfrac{\pi}{2}} &= \begin{pmatrix}
2 \sin 4 \theta_2 & -4 \sin 4 \theta_2 & -2 \sin 4 \theta 2
  \end{pmatrix}\\
H(C_{L_4})|_{\theta_1, \theta_3=\tfrac{\pi}{2}} & =\begin{pmatrix}
-8 \cos 4 \theta_2 & 8 \cos 4 \theta_2 & 4 (\cos 2 \theta_2 + \cos 4 \theta_2) \\
8 \cos 4 \theta_2 & -16 \cos 4 \theta_2 & -8 \cos 4 \theta_2 \\
4 (\cos 2 \theta_2 + \cos 4 \theta_2) & -8 \cos 4 \theta_2 & -8 \cos 4 \theta_2
  \end{pmatrix}\\
\text{EVal.}(H_{L_4})|_{\theta_1, \theta_3=\tfrac{\pi}{2}} & =\begin{pmatrix}
4 (\cos 2 \theta_2 - \cos 4 \theta_2) \\
-2 \cos 2 \theta_2 - 14 \cos 4 \theta_2 - \ldots\\
\ldots \sqrt{2} \sqrt{34 - 2 \cos 2 \theta_2 + \cos 4 \theta_2 - 2 \cos 6 \theta_2 + 33 \cos 8 \theta_2} \\
 -2 \cos 2 \theta_2 - 14 \cos 4 \theta_2 + \ldots \\
 \ldots \sqrt{2} \sqrt{34 - 2 \cos 2 \theta_2 + \cos 4 \theta_2 - 2 \cos 6 \theta_2 + 33 \cos 8 \theta_2}
  \end{pmatrix}\\
\text{EVec.}(H_{L_4})|_{\theta_1, \theta_3=\tfrac{\pi}{2}} & =
\begin{pmatrix}
1 \\
0 \\
1 
 \end{pmatrix},\\
&\begin{pmatrix}
-1 \\
 (\frac{\sqrt{2}}{8} \sqrt{\begin{matrix}
2 \cos 2 \theta_2+\cos 4 \theta_2- \ldots\\
\ldots 2 \cos 6 \theta_2 +33 \cos 8 \theta_2+34
\end{matrix}-}\ldots\\
\ldots -2 \cos 2 \theta_2) \sec 4 \theta_2+\frac{1}{4} \\
1 
\end{pmatrix},\\
&\begin{pmatrix}
-1 \\
\frac{1}{4}-(2 \cos \frac{1}{4} \theta_2+\ldots \\
\ldots \frac{\sqrt{2}}{8} \sqrt{\begin{matrix}
-2 \cos 2 \theta_2+\cos 4 \theta_2-2 \cos 6 \theta_2+ \ldots\\
\ldots 33 \cos 8 \theta_2+34
\end{matrix}})\sec 4 \theta_2 \\
1 
\end{pmatrix}
\end{split}
\end{equation}

\section{Proofs of Theorems~\ref{thm:globalminimum} and~\ref{thm:rotations}}
\renewcommand{\thefigure}{B\arabic{figure}}

\setcounter{figure}{0}
\subsection{$L_2$ cost minima and equiangular tight-frames: proof of Theorem~\ref{thm:globalminimum}}\label{proof1}

Here we prove Theorem~\ref{thm:globalminimum} in two steps: first we can show the equivalence, up to an additive constant, of minimizing the $L_2$ cost and minimizing the $L_2$ norm of the error of Eq~\ref{eq:outer}. Then we show that the pathological solution (Section~\ref{sec:l2}) is at the global minimum of this cost.

\begin{proof}[Proof of Theorem~\ref{thm:globalminimum}]
For a normalized ($\sum_kW_{ik}^2=1,\ \forall\ i$) matrix, $W$:
\begin{equation}
\begin{split}
C_{L_2} &= \sum_{ij}(\sum_k W_{ik}W_{jk}-\delta_{ij})^2\\
&= \sum_{ij}(\sum_k W_{ik}W_{jk}-\delta_{ij})(\sum_l W_{il}W_{jl}-\delta_{ij})\\
&= \sum_{ijkl}W_{ik}W_{jk}W_{il}W_{jl}-2\sum_{ijk}W_{ik}W_{jk}\delta_{ij}+\sum_{ij}\delta_{ij}^2\\
&= \sum_{ijkl}W_{ik}W_{jk}W_{il}W_{jl}-2\sum_{ik}W_{ik}^2+\textrm{const.}(L)\\
&= \sum_{ijkl}W_{ik}W_{jk}W_{il}W_{jl}+\textrm{const.}(L)\\
\end{split}
\end{equation}

\begin{equation}
\begin{split}
C_{\text{Eq}~\ref{eq:outer}} &= \sum_{kl}(\sum_i W_{ik}W_{il}-\frac{L}{D}\delta_{kl})^2\\
&= \sum_{kl}(\sum_i W_{ik}W_{il}-\frac{L}{D}\delta_{kl})(\sum_j W_{jk}W_{jl}-\frac{L}{D}\delta_{kl})\\
&= \sum_{ijkl}W_{ik}W_{il}W_{jk}W_{jl}
-2\sum_{ikl} \frac{L}{D}W_{ik}W_{il}\delta_{kl}
+\sum_{kl}(\frac{L}{D}\delta_{kl})^2\\
&= \sum_{ijkl}W_{ik}W_{il}W_{jk}W_{jl}
-2\frac{L}{D}\sum_{ik} W_{ik}^2+\textrm{const.}(L, D)\\
&= \sum_{ijkl}W_{ik}W_{il}W_{jk}W_{jl}+\textrm{const.}(L, D)\\
\end{split}
\end{equation}

where $\sum_kW_{ik}^2=1,\ \forall\ i$ is used extensively and the index letters were initially chosen to make the comparison of the final lines more clear. In~\cite{Le2011}, this first equivalence was shown and it was also shown that the $L_2$ cost is equivalent to the reconstruction cost with whitened data (Lemmas 3.1 and 3.2).

Now we can show that the same dictionary that was described in Section~\ref{sec:l2}: $W_0$, an integer overcomplete dictionary where each set of complete bases is an orthonormal basis, exactly satisfies Eq~\ref{eq:outer} and so is a minimum of the $L_2$ cost. This solution is very far away from an ETF in the sense of Eq~\ref{eq:equiangular}. A dictionary of this form, $W\in \mathbb{R}^{L\times D}$, can be constructed as $W_{ij} = \delta_{(i\ \textrm{mod}\ D)j}$ with $L=n\times D,\ n>1,\ \in\mathbb{Z}$, that is, a $D$ dimensional identity matrix tiled $n$ times.

This construction satisfies Eq~\ref{eq:outer} and therefore has a value of 0 for $C_{\text{Eq}~\ref{eq:outer}}$. Since $C_{\text{Eq}~\ref{eq:outer}}$ is a sum of quadratic, and therefore non-negative, terms, this construction is a global minimum of $C_{\text{Eq}~\ref{eq:outer}}$ and the $L_2$ cost.
\begin{equation}
\begin{split}
\sum_kW_{ki}W_{kj} &= \sum_k \delta_{(k\ \textrm{mod}\ D)i}\delta_{(k\ \textrm{mod}\ D)j}\\
&=n\delta_{ij}\\
&=\frac{L}{D}\delta_{ij}\\
&\Rightarrow C_{\text{Eq}~\ref{eq:outer}}=0
\end{split}
\end{equation}
as $k\ \textrm{mod}\ D=i$ a total of $n$ times when $i=j$.

However, this construction has off-diagonal Gram matrix elements that are either 0 or 1:
\begin{equation}
\begin{split}
\cos \theta_{ij} &=\sum_kW_{ik}W_{jk}\\
&=\sum_k \delta_{(i\ \textrm{mod}\ D)k}\delta_{(j\ \textrm{mod}\ D)k}\\
&=\delta_{(i\ \textrm{mod}\ D)(j\ \textrm{mod}\ D)},
\end{split}
\end{equation}
which is not equal or close to an equiangular solution, that is, $\cos\theta_{ij}=\cos\alpha,\ \forall i\neq j$.
\end{proof}

\subsection{Invariance to continuous transformations: proof of Theorem~\ref{thm:rotations}}\label{proof2}

Here we prove Theorem~\ref{thm:rotations}: the $L_2$ cost, initialized from the pathological solution, is invariant to transformations, $\Phi$, constructed as orthogonal rotations applied to any basis subset and an identity transformation on the remaining bases. This shows that low coherence and high coherence configurations are both global minima of the $L_2$ cost.

\begin{proof}[Proof of Theorem~\ref{thm:rotations}]
For an $D$ dimensional space with an $n$ times overcomplete dictionary, with $n$ an integer greater than 1, the pathological dictionary configuration is a orthonormal basis tiled $n$ times. The dictionary elements can be labels as the sequential subsets of orthornormal subsets $W_1,\dots,W_D,\ldots,W_{2D},\ldots,W_{n\!\times\! D}$. So, bases $W_1$ through $W_D$ form a full-rank, orthonormal basis and this basis is tiled $n$ times.

Consider the following partition of the bases: partition $\mathcal A$ is the first orthonormal set, bases $W_1$ through $W_D$, and partition $\mathcal B$ the remainder of the bases, $W_{D\!+\! 1}$ through $W_{n\!\times\!D}$. Let $P$ be a projection operator for $\mathcal A$ and $P^C$ its compliment projection operator, that is, $P^CW_i=W_i$ and $PW_i=0\ \forall\ W_i \in \mathcal B$ and $PW_j=W_j$ and $P^CW_j=0\ \forall\ W_j \in \mathcal A$. Let $R\in \text{O}(L)$ be a rotation and $PR$ a rotation that only acts on the $\mathcal A$ subspace. The operator $\Phi=PR + P^C$ is a rotation applied to all elements of $\mathcal A$ which leaves elements of $\mathcal B$ unchanged. Under its action, only terms in the cost between elements of $\mathcal A$ and $\mathcal B$ will change. It is straightforward to show that the terms in the cost that have both elements within $\mathcal A$ or both within $\mathcal B$ are constant since the rotation does not alter the relative pairwise angles.

For $W_i \in \mathcal B$, we can write down the terms in the $L_2$ cost which contain itself and elements from $\mathcal A PR$:
\begin{equation}
\begin{split}
C_{W_i}(\mathcal A\Phi) &= \sum_{W_j \in \mathcal A}(R^TP^TW_j^TW_i)^2+(W_i^T W_jPR)^2\\
&= \sum_{W_j \in \mathcal A}(R^TW_j^TW_i)^2+(W_i^T W_jR)^2\\
&=2\sum_{W_j \in \mathcal A}\text{Proj}_{ W_jR}(W_i)^2\\
&=2|W_i|^2\\
&=C_{W_i}(\mathcal A).
\end{split}
\end{equation}

Since the $W_j\in \mathcal A$  remain an orthonormal basis under a rotation, the sum of the projections-squared is the $L_2$ norm-squared of $W_i$ which is constant. Since this is true for every $W_i \in \mathcal B$, the entire cost is constant under this transformation. This argument holds for any subset which forms an orthonormal basis and so all orthonormal subsets can rotate arbitrarily with respect to each other without changing the value of the $L_2$ cost, but the coherence of the matrix does depend on the transformation, $\Phi$. This shows that the $L_2$ global minimum contains dictionaries with coherence $=1$ and $<1$ which can be continuously transformed into each other.
\end{proof}

\section{Additional figures}
\renewcommand{\thefigure}{C\arabic{figure}}

\setcounter{figure}{0}
\subsection{Eigenvalues of the $L_2$ and $L_4$ cost in a 2 dimensional, 2 times overcomplete example}

\begin{figure}[!htbp]
  \centering
   \includegraphics{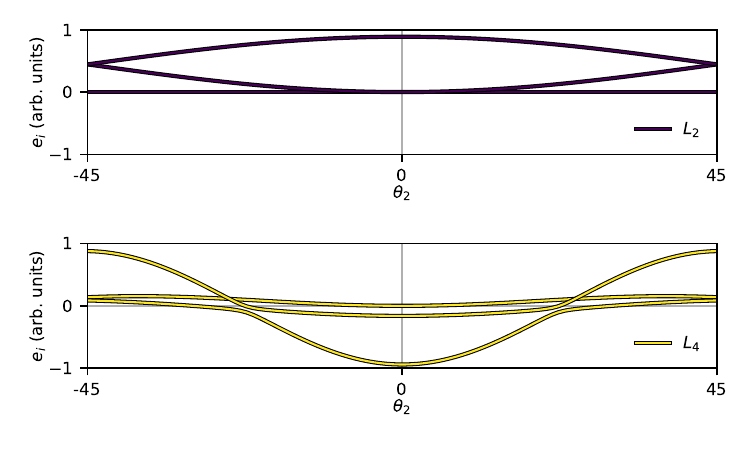}
  \caption{Eigenvalues of the Hessian for the $L_2$ and $L_4$ costs for a pair of orthonormal bases as a function of the angle between the pairs. \textbf{A} The eigenvalues of the Hessian of the $L_2$ cost evaluated at $\theta_1 = \theta_3 = \pi/2$ as a function of $\theta_2$. Each purple line is one of the three eigenvalues of the Hessian of the $L_2$ cost as $\theta_2$ is varied. \textbf{B} Same as \textbf{A} but for the $L_4$ cost. Each yellow line is one of the three eigenvalues of the Hessian of the $L_4$ cost as $\theta_2$ is varied.}
  \label{fig:eig}
\end{figure}

Fig~\ref{fig:eig} shows all eigenvalues for the $L_2$ and $L_4$ costs for the 2 dimensional problem in Section~\ref{sec:l2}.

\subsection{Extended Fig~\ref{fig:simulations}}
\label{S5_Figure}
\begin{figure}[!htbp]
  \centering
   \includegraphics{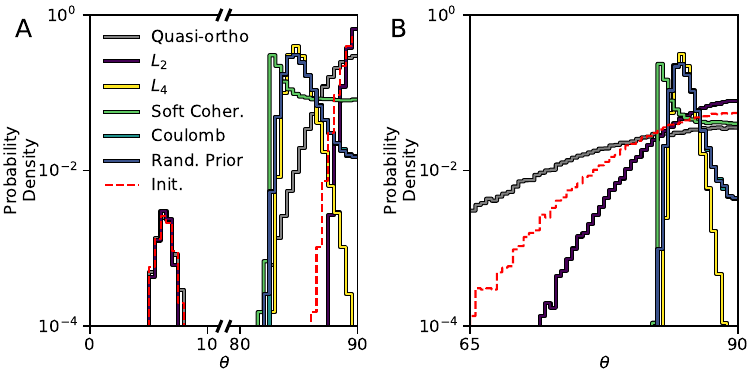}
  \caption{Coherence control costs have minima with varying coherence which can depend on initialization. Color legend is preserved across panels. For both panels a 2 times overcomplete dictionary with a data dimension of 64 was used. \textbf{A} Distribution of pairwise angles (log scale) obtained by numerically minimizing a subset of the coherence cost functions for the pathological dictionary initialization. Red dotted line indicates the initial distribution of pairwise angles. Note that the horizontal axis is broken at 10 and 80 degrees. \textbf{B} Angle distributions obtained (as in \textbf{A}) from a uniform random dictionary initialization. Note that the horizontal axis only includes 65 to 90 degrees.}
  \label{fig:simulations_all}
\end{figure}

{\bf }
Fig~\ref{fig:simulations_all} is identical analysis as Fig~\ref{fig:simulations} with all cost functions included.

\subsection{Extended Fig~\ref{fig:recovery}}
\begin{figure}[!htbp]
  \centering
   \includegraphics{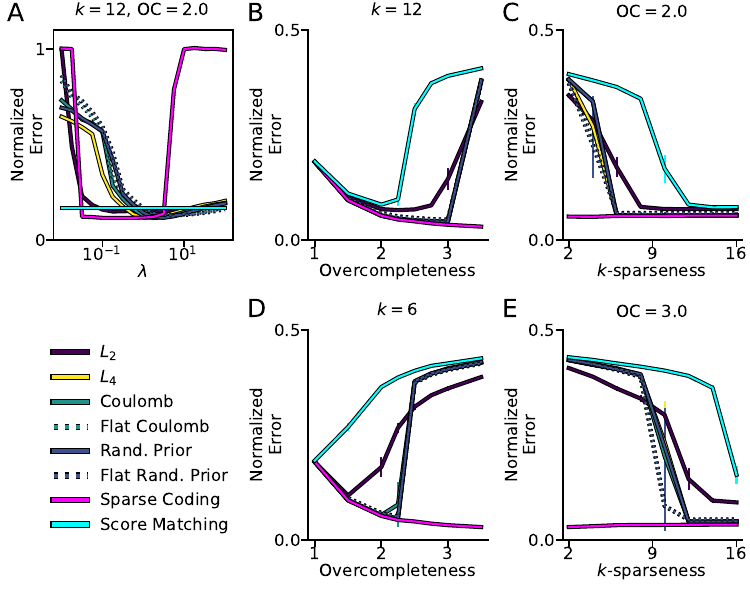}
  \caption{Coherence control costs do not all recover mixing matrices well. All ground truth mixing matrices were generated from the Soft Coherence cost and had a data dimension of 32. Color and line style legend are preserved across panels. \textbf{A} The normalized recovery error (see Section~\ref{methods} for details) for a 2-times overcomplete mixing matrix and $k=12$ as a function of the sparsity prior weight ($\lambda$).  Since score matching does not have a $\lambda$ parameter, it is plotted at a constant. \textbf{B} Recovery performance ($\pm$ s.e.m., $n=10$) at the best value of $\lambda$ as a function of overcompleteness at $k=12$. \textbf{C} Recovery performance ($\pm$ s.e.m., $n=10$) at the best value of $\lambda$ as a function of $k$-sparseness at 2-times overcompleteness. \textbf{D, E} Same plots as \textbf{B} and \textbf{C} at a point where methods do not perform as well: $k=6$ and 3-times overcomplete.}
  \label{fig:recovery_all}
\end{figure}

Fig~\ref{fig:recovery_all} is identical analysis as Fig~\ref{fig:recovery} with all cost functions included.

\subsection{Supplemented Fig~\ref{fig:flat}D.}
\begin{figure}[!htbp]
  \centering
   \includegraphics{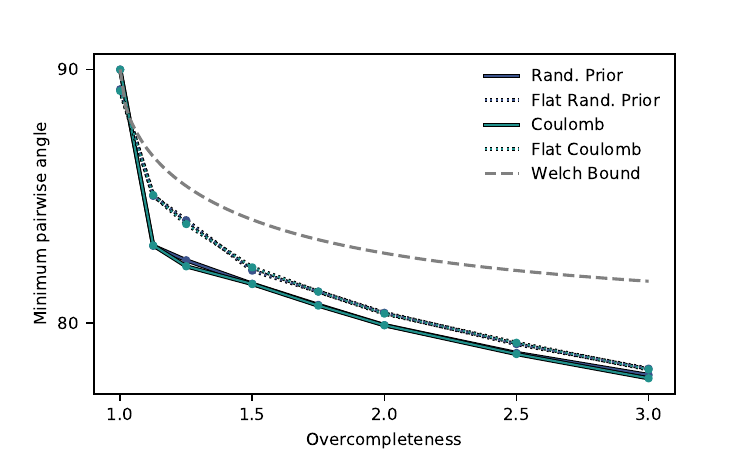}
  \caption{Quadratic terms dominate the minima of coherence control costs as a function of overcompleteness. The median minimum pairwise angle (arccosine of coherence) across 10 initializations is plotted as a function of overcompleteness for a dictionary with a data dimension of 32. The largest possible value (Welch Bound) is also shown as a function of overcompleteness.}
  \label{fig:3D_all}
\end{figure}

Fig~\ref{fig:3D_all} is similar to Fig~\ref{fig:flat}D for the Coulomb and Random Prior costs and their flattened versions.

\subsection{Comparison to V1 Receptive fields}
Comparison of $n_x=\frac{\sigma_x\ k}{2\pi}$ and $n_y=\frac{\sigma_y k}{2\pi}$, the number of wavelengths in the $x$ and $y$ directions contained in the Gabor envelope to data collected from macaque V1~\cite{Ringach2002}.

\begin{figure}[!htbp]
  \centering
   \includegraphics{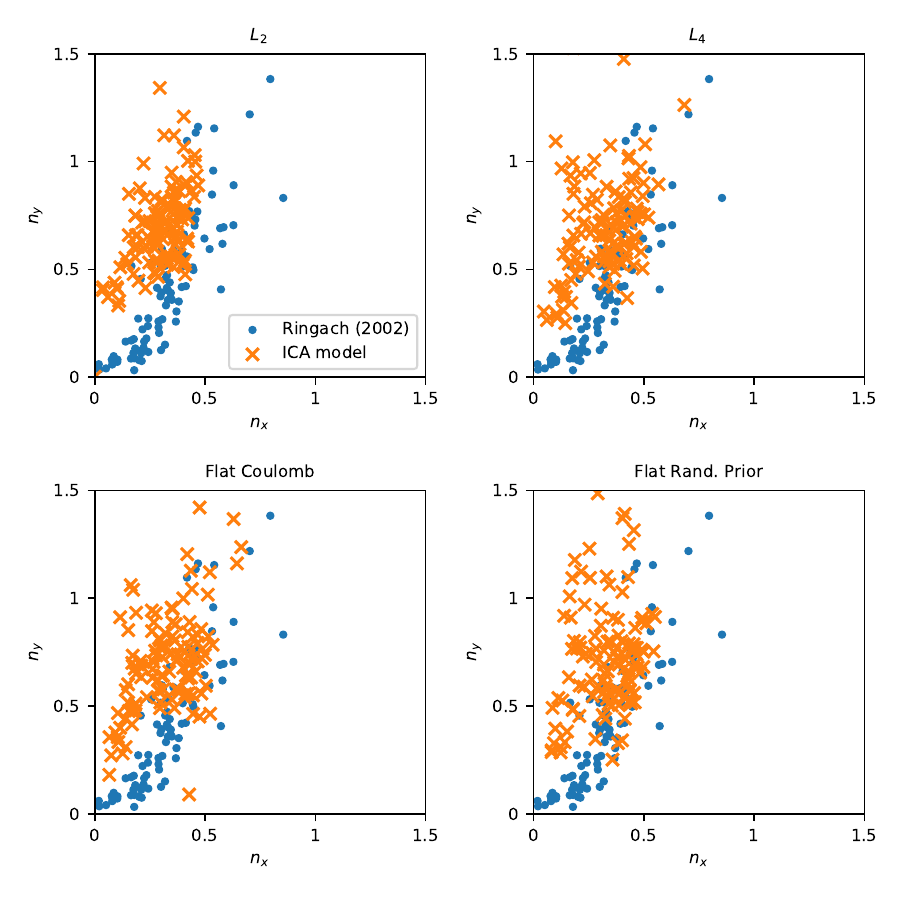}
  \caption{2-times overcomplete ICA models do not match the distribution of simple cell receptive fields reported in macaque V1 unlike sparse coding approached designed to better mimic biological neurons~\cite{Rehn2007, Zylberberg2011}.}
  \label{fig:v1_comparison}
\end{figure}

\end{document}